\documentclass[twoside,11pt]{article}

\usepackage{mathtools}
\usepackage{jmlr2e}
\usepackage{natbib} 
\usepackage{hyperref} 
\usepackage{cleveref} 

\DeclareMathAlphabet{\mathpzc}{OT1}{pzc}{m}{it}

\ShortHeadings{Generalized Laguerre Reduction of Volterra Kernel}{Israelsen and Smith}
\firstpageno{1}

\begin{document}

\title{Generalized Laguerre Reduction of the Volterra Kernel for Practical Identification of Nonlinear Dynamic Systems}

\author{\name Brett W. Israelsen \email brett.israelsen@gmail.com
       \AND
       \name Dale A. Smith\email dale.smith@apco-inc.com \\
       \addr Process Control Engineering\\
       Advanced Process Control and Optimization Inc. (APCO Inc.)\\
       Salt Lake City, UT 84116, USA}

\editor{TBD}

\maketitle

\begin{abstract}
The Volterra series can be used to model a large subset of nonlinear, dynamic systems. A major drawback is the number of coefficients required model such systems. In order to reduce the number of required coefficients, Laguerre polynomials are used to estimate the Volterra kernels. Existing literature proposes algorithms for a fixed number of Volterra kernels, and Laguerre series. This paper presents a novel algorithm for generalized calculation of the finite order Volterra-Laguerre (VL) series for a MIMO system. An example addresses the utility of the algorithm in practical application. 

\end{abstract}


\section{Introduction}
\label{sec:VolterraSeriesBackground}
The Volterra Series were first studied by Vito Volterra and were named after him. The first application of the Volterra series to the study of nonlinear systems was done by Norbert Wiener \citep[see][p.517]{Schetzen1980}. The time invariant series can be represented by \eqref{eq:GenVoltConst} below.

\begin{align}
y\left ( t \right )=&\int_{-\infty}^{\infty }h_1(\sigma_1)u(t-\sigma_1)d\sigma_1 \nonumber \\
+&\int_{-\infty}^{\infty }\int_{-\infty}^{\infty }h_2(\sigma_1,\sigma_2)u(t-\sigma_1)u(t-\sigma_2)d\sigma_1d\sigma_2 + \cdots \nonumber\\ 
+&\int_{-\infty}^{\infty }\cdots \int_{-\infty}^{\infty }h_N(\sigma_1,\cdots,\sigma_i)u(t-\sigma_1)\cdots \nonumber\\
&\qquad \qquad \qquad u(t-\sigma_N)d\sigma1\cdots d\sigma_N \label{eq:GenVoltConst}
\end{align}

Here $u$ represents the input to the system, $y$ is the system output and $h_n$ is known as the $n^{th}$ Volterra kernel, it can also be called the $n^{th}$ order impulse response. This terminology comes because the first term of \cref{eq:GenVoltConst} is the same as the convolution integral which relates the output of a system to the input and the $1^{st}$ order impulse response of the system\footnote{Note that \cref{eq:GenVoltConst}, and consequently most of the other equations in this paper represent\ SISO systems for compactness. MIMO systems can be accounted for by solving several MISO sets, and expanding the expression with $u_{i}(t) \text{, }i=1 \ldots I$ where $I$ is the total number of inputs. For illustrative purposes it is sufficient to look at SISO systems for now.}. Higher order terms of the Volterra series can be seen as higher order impulse responses. The terms \emph{Volterra kernel} and \emph{impulse response} will be used interchangeably in this document.

Several modifications can and should be made to \eqref{eq:GenVoltConst} before performing an identification. These modifications include: discretization of the integrals, modification of the limits of integration based on known characteristics of physical systems, and reduction of the Volterra kernel using the Laguerre polynomials to reduce the number of model parameters.

\section{Simplifying the Volterra Series for Practical Application}
If a system is \emph{causal}, which means that the output at some time $t$ depends only on past inputs($u(t-\sigma)$ for $\sigma>0$) and not on future inputs ($u(t-\sigma)$ for $\sigma<0$). The Volterra series can be written as shown in \cref{eq:GenVoltCausal} below. Note that \eqref{eq:GenVoltCausal} only includes the first term of the series for simplicity. It should also be noted that all known physical systems appear to be \emph{causal}. A more detailed discussion of \emph{causality} can be found in \citet[p.21,89]{Schetzen1980}.

\begin{align}
    y\left ( t \right )=&\int_{0}^{\infty }h_1(\sigma_1)u(t-\sigma_1)d\sigma_1
\label{eq:GenVoltCausal}
\end{align}

The Volterra series may also be discretized by using the convolution sum instead of the convolution integral, yielding \cref{eq:GenVoltDisc}. Again, only the first term of the series is shown for simplicity.

\begin{align}
    y\left ( t \right )=\sum_{i_1=0}^{\infty }h_1(i_1)u(t-i_1)
\label{eq:GenVoltDisc}
\end{align}

Finally, if the system is assumed to have \emph{finite memory} or \emph{fading memory} and finite order another simplification can be made. \emph{Fading Memory} means that the there is some time $M$ in the past before which inputs will no longer have affect on the output of the system.  \Cref{eq:GenVoltDiscFiniteMem} is referred to as the discrete, finite memory, $N^{th}$ order Volterra Series. 

\begin{align}
y\left ( t \right )=&\sum_{n=1}^{N} \nu_{M}^{n}(t) \label{eq:GenVoltDiscFiniteMem}\\
\nu_{M}^{n}(t) =&\sum_{i_1=0}^{M}\cdots \sum_{i_n=0}^{M}h_n(i,\ldots,i_n)u(t-i_1)\cdots u(t-i_n) \nonumber
\end{align}

\subsection{Volterra Model Limitations}
This class of finite Volterra models is defined as the class of $V_{(N,M)}$ models by \citet{DoyleOgunnaikePearson2001}. Where $N$ is the nonlinear degree and $M$ is the dynamic order. In other words $N$ is the number of Volterra terms and $M$ is the memory length of the system. Using this notation it is easy to describe different Volterra Models by examining the behaviors of the limiting cases. These are: $V_{(\infty,M)}$,$V_{(N,\infty)}$, and $V_{(\infty,\infty)}$ \citep[Ch. 2]{DoyleOgunnaikePearson2001}, \citep[Sec.4.2]{Pearson1999}.

It is important to consider the limitations of the $V_{(N,M)}$ class. Some of these limitations include not being able to exhibit \emph{output multiplicity} \citep{boyd1985fading}. This can be described intuitively by saying that if a system can exhibit the same output by different local inputs (i.e. different steady state responses to the same steady state input), it must have had paths that differed initially. This leads to a similar conclusion which says conditionally stable impulse responses cannot be described by a \emph{fading memory} Volterra model. Volterra models also cannot produce persistent oscillations or chaos in response to asymptotically constant input sequences.

The limitations of the Volterra series can be seen as beneficial or detrimental depending on the desired output of the model. If a model for a system with persistent oscillations is desired then VL models should not be used. However, it is useful to have a model that implicitly rejects these types of characteristics if the physical system does not exhibit them. More information on this subject can be found in \citet{DoyleOgunnaikePearson2001,Pearson1999}

\subsection{Volterra Model Parameterization}\label{sec:VoltParameterization}
Another important practical limitation, that isn't dependent on the application, is \emph{Volterra Model Parameterization}. In other words how many parameters are required to define a $V_{(N,M)}$ model. The total number of parameters can be represented as $C_{(N,M)}$ \citep{DoyleOgunnaikePearson2001}. The following equations describe the calculation of $C_{(N,M)}$.

\begin{align}
C_{(N,M)} &= \sum_{n=0}^{M} C_n(M) \\
C_n(M) &= (M+1)^n \nonumber\\
C_{(N,M)} &= \sum_{n=0}^{M} C_n(M) = \frac{(M+1)^{N+1}-1}{M}\label{eq:VoltParam} \\
&\simeq M^N \nonumber
\end{align}

Here $C_n(M)$ is the total number of coefficients in $h_n$(The $n^{th}$ Volterra Kernel) of the Volterra model $V_{(N,M)}$. Although \citet[p.19]{DoyleOgunnaikePearson2001} discusses methods for reducing the number of coefficients, the relationship is still exponential and therefore remains a barrier for practical application. \Cref{tab:VolterraParameterization} below demonstrates how quickly the number of required parameters can grow, especially considering that $M$ is regularly between $50$ and $250$ in many industrial processes. The number of model parameters required makes any Voterra model with $N > 2$ impratical.

\begin{table}[htb]
\centering
\begin{tabular}{|r|r|r|r|r|}
\hline
   & N=1 & N=2 & N=3 & N=4 \\ \hline
M=1 & 1   &  1  &  1  &  1  \\ \hline
M=2 & 2   &  4  &  8  &  16 \\ \hline
M=3 & 3   &  9	&  27 &  81 \\ \hline
M=4 & 4   &  16 &  64 &  256\\ \hline
M=10& 10  &  100&  1000& 10\,000\\ \hline
M=20& 20  &  400&  8000& 16\,000\\ \hline
M=50& 50  &  2500& 125\,000& 6\,250\,000\\ \hline 
\end{tabular}
\caption{Number of Volterra Parameters based on N and M}
\label{tab:VolterraParameterization}
\end{table}

\section{The Laguerre Polynomials}\label{sec:LaguerrePolynomialsBackground}
Laguerre Polynomials are named after Edmond Laguerre. The Laguerre Polynomials are a series of orthogonal polynomials that can be used to reduce the number of coefficients required to describe a Volterra kernel. The application of orthogonal functions to identification and control is not new, see \citep{Schetzen1980}, \citep{dumont1991}, \citep{zheng1994}, \citep{makila1990approximation}, \citep{clement1982laguerre}, \citep{Mahmoodi2007}, \citep{Zheng1995},and \citep{CommunicationLee1960}. A mathematical review of orthogonal and orthonormal fucntions can be found in Appendix \ref{sec:Orthogonality}.

\subsection{Making the Laguerre Functions}
\subsubsection{Forming the General Laguerre Representation}
The Laguerre functions can be obtained by forming an orthonormal set from the linearly independent set of functions in \eqref{eq:RootLag}. Note that since the functions are only non-zero for $t\geq 0$ that the Laguerre functions will be orthonormal on the domain $[0,\infty)$.

\begin{align}
v_n = 
\begin{cases}
0 &\text{for }t<0\\
(at)^ne^{-at} &\text{for }t\geq 0; n=0,1,2,\ldots \label{eq:RootLag}
\end{cases}
\end{align}

The Laguerre function can then be represented as:
\begin{equation}
l_n(t)=\sum_{m=0}^{\infty}c_{mn}v_m(t) \label{eq:LagDef}
\end{equation}

It must adhere to the orthonormal condition given in Equations \eqref{eq:OrthonormCond},\eqref{eq:OrthonormCond_a} and \eqref{eq:OrthonormCond_b} (see Appendix \ref{sec:Orthogonality}). This can be done by choosing the coefficients $c_{mn}$ in order to satisfy the given equations. Examples of this procedure for the time and frequency domain can be found in \citet[Sec. 16.1-16.2]{Schetzen1980}. It can be shown that the general expression for the Laguerre functions in the time domain is \cref{eq:GeneralLaguerre} below.

\begin{align}
l_n(t) = \sqrt{2a}\sum_{k=0}^{n}\frac{(-1)^{k}n!2^{n-k}}{k![(n-k)!]^2}(2at)^{n-k}e^{-at}
\label{eq:GeneralLaguerre}
\end{align}

\subsubsection{Laguerre Time Scale Factor -- $a$}\label{sec:LaguerreTimeScale}
There is some interesting discussion that should take place concerning the value $a$ in \cref{eq:GeneralLaguerre}. It is a factor by which the time scale of the Laguerre functions can be lengthened or shortened. It is often referred to as the \emph{Laguerre time scale} because of this. Examination of the frequency domain representation of the Laguerre polynomials shows that the $a$ term defines the time constant of a filter. This leads to more discussion on the use of Laguerre polynomials as filters. It is sufficient to know for the purposes of this paper that the parameter $a$ or rather $1/a$ is the time constant of the ``Laguerre Filter" and that because of this filter the Laguerre functions can reject noise in experimental data if it is chosen properly. 

\Cref{eq:GenLagFreq,,eq:GenLagFreqFilt} below are frequency domain representations of the Laguerre polynomials and make the filter more obvious. Observe that $\alpha$ is a low pass filter and $\beta$ is an all-pass filter making the net effect of the Laguerre Functions a filter with time constant $1/a$. For more detail on the frequency domain representation of the Laguerre Functions see \citet[Sec. 16.2]{Schetzen1980}. For more information on the Laguerre filter see \citet{Silva1995} and \cite{king1977}.

\begin{align}
L_n(s)&=\sqrt{2a}\frac{(a-s)^n}{(a+s)^{n+1}}; \sigma > -a \label{eq:GenLagFreq}\\
&\text{or} \nonumber \\
L_n(s)&=\underbrace{\left[\frac{\sqrt{2a}}{a+s}\right]}_{\alpha}\underbrace{\left[\frac{(a-s)}{(a+s)}\right]^n}_{\beta}; \sigma > -a \label{eq:GenLagFreqFilt}
\end{align}

Research has been performed to identify the optimal time scale factor $a$ for a given identification problem. It has been suggested that the factor should be placed near the dominant pole of the system \citep{Zheng1995}, however if the system has delay the factor $a$ will be greatly affected \citep{wang1994optimal}. Methods exist for calculating the optimal time scale value for linear systems, or require previous knowledge of the system and thus are not generally applicable to nonlinear system identification \citep{Clowes1965}, \citep{Fu1993}, and \citep{parks1971}. Current general practice is to perform a nonlinear optimization to calculate the value for $a$ that will yield the minimum error.

If the value for $a$ is optimal the coefficients of the higher order terms of the Laguerre polynomial will go to zero. This is valuable because the main purpose of using the Laguerre functions is to reduce the number of parameters that need to be identified for a $V_{(N,M)}$ model. If the value of $a$ is not optimal then the Laguerre functions can still be used but a higher order Laguerre polynomial will be required. Some further discussion of properties can be found in Appendix \ref{sec:LaguerreProps}.

\section{Laguerre Estimation of the Volterra Kernel}
Recall the first order discrete Volterra kernel given in \cref{eq:GenVoltDisc} (shown below for reference).
\begin{align}
    y\left ( t \right )=\sum_{i_1=0}^{\infty }h_1(i_1)u(t-i_1) \nonumber
\end{align}

The only unknown is $h_1(i_1)$, the $1^{st}$ order impulse response, since for system identification both $y(t)$ and $u(t)$ are recorded I/O data. $h_1(i_1)$ can be approximated by linear combination of the Laguerre functions. 

In the case of the first order Volterra-Laguerre series. The first order Volterra kernel ($h_1(i_1)$) is approximated by linear combination of an $r^{th}$ order Laguerre polynomial. $h_1(i_1)$ meets the requirement of its square being finite over the interval through which the Laguerre functions are orthonormal (see \eqref{eq:ISErestrict}).The formulation is shown below:

\begin{align}
h_1(i_1)\approx \sum_{r=1}^{R} \theta_rl_r(t)
\label{eq:VoltLag1OrderKernel}
\end{align}

Here, $l_r(t)$ is given by \eqref{eq:GeneralLaguerre} and is shown below for reference.

\begin{align}
l_r(t) = \sqrt{2a}\sum_{k=0}^{r}\frac{(-1)^{k}r!2^{r-k}}{k![(r-k)!]^2}(2at)^{r-k}e^{-at} \nonumber
\end{align}

Substituting \cref{eq:VoltLag1OrderKernel} into \cref{eq:OrthogApprox} and truncating both $h_1(i_1)$ and $l_r(t)$ to a memory length of $M$ yields:

\begin{align}
y(t)\approx \sum_{i_1=0}^{M}\sum_{r=1}^{R}\theta_rl_r(t)u(t-i_1)
\end{align}

Now, defining the following:
\begin{align}
\mathbf{\Theta} &= [\theta_1,\theta_2,\ldots,\theta_R]^T \\
\mathbf{B} &= 
\begin{bmatrix}
l_1(0) & l_2(0) &\cdots &l_R(0) \\
l_1(1) & l_2(1) &\cdots &l_R(1) \\
\vdots & \vdots &\ddots &\vdots \\
l_1(M) & l_2(M) &\cdots &l_R(M) \\
\end{bmatrix} \\
\mathbf{U_k}&=[u(k),u(k-1),\ldots,u(k-M)] 
\end{align}

Then the Volterra system can be approximated by:
\begin{align}
\tilde{y}(k) = \mathbf{U}_k\mathbf{B\Theta}
\end{align}

In order to extend the representation to higher order Volterra series for a MIMO system it is first useful to define the reduced Kronecker product as \eqref{eq:RKronecker} \citep[p.100]{Rugh1981} :

\begin{align}
a^{[2]}=a\otimes a=&[a_1,a_2,\ldots,a_n]^{[2]}= \nonumber \\
&[a_1a_1,a_1a_2,\ldots,a_2a_2,a_2a_3,\ldots,a_na_n] \label{eq:RKronecker}
\end{align}

Using the reduced Kronecker product notation above a general MISO Volterra-Laguerre series with $I$ inputs can be approximated by \eqref{eq:MISOMatrix} below. For a MIMO system the separate MISO solutions can be combined.

\begin{align}
\tilde{y}(k) &= [\mathbf{U}_k,\mathbf{U}_k^{[2]},\ldots,\mathbf{U}_k^{[N]}]\mathbf{\Theta} \label{eq:MISOMatrix}\\
\text{where} \nonumber \\
\mathbf{U}_k^i &= [u_i(k),\ldots,u_i(k-m)],  i=1,\ldots,I \\
\mathbf{U}_k &=  [\mathbf{U}_k^1\mathbf{B},\ldots,\mathbf{U}_k^I\mathbf{B}]
\end{align}

This notation was originally derived in \citet{zheng2004volterra}.

\section{Added generalization for Ease of Practical Application}
In order to simplify practical application, generalizations were made to make the algorithm more easily scalable. Each of the generalizations will be discussed separately and are listed below:

\begin{enumerate}
\item Allow different nonlinear degree $N$ for each system input $i=1 \ldots I$
\item Allow different Laguerre series $l_{r}$ for each Volterra term $n=1 \ldots N \text{ and input } i=1 \ldots I$
\item Allow different Laguerre time scale $a_{n,i}$ for each Laguerre series $l_{r}^{a_{n,i}}$
\end{enumerate}

\subsection{Separate Volterra order for each system input}
The Volterra series can be used to model a large class of systems.\Cref{eq:GenVoltConst} (shown below for reference) is the general $N^{th}$ order Volterra series. 

\begin{align}
y\left ( t \right )=&\int_{-\infty}^{\infty }h_1(\sigma_1)u(t-\sigma_1)d\sigma_1 \nonumber \\
+&\int_{-\infty}^{\infty }\int_{-\infty}^{\infty }h_2(\sigma_1,\sigma_2)u(t-\sigma_1)u(t-\sigma_2)d\sigma_1d\sigma_2 + \cdots \nonumber\\ 
+&\int_{-\infty}^{\infty }\cdots \int_{-\infty}^{\infty }h_N(\sigma_1,\cdots,\sigma_i)u(t-\sigma_1)\cdots \nonumber\\
&\qquad \qquad \qquad u(t-\sigma_N)d\sigma1\cdots d\sigma_N\nonumber
\end{align}

$N=1$ is an example of the first order Volterra series and can describe systems with a linear relationship between the inputs and output. Using $N=2$ allows the Volterra series to describe second order relationships between the inputs and output. It cannot be assumed that, in a general system all inputs will have the same relationship to the output. In fact this will rarely be the case.

\subsection{Separate Laguerre series for each Volterra term}
Considering a single input being mapped to an output using an $N^{th}$ order Volterra series. There will be $N$ different Laguerre polynomials to approximate each kernel of the Volterra series. Depending on the complexity of the relationship between one term of the Volterra series and the output; a different order $R$ of the Laguerre polynomial could be used for each separate term of the Volterra series. This option gives one the ability to use more or less Laguerre coefficients to aproximate a certain term of the Volterra series if needed. A general example of an input with an uncomplicated first order relationship to the output and a more complicated second order relationship to the output could take advantage of fewer laguerre polynomials to approximate the first order Volterra kernel and more laguerre polynomials to approximate the second order Volterra kernel. Myriad other scenarios exist where this flexibility would be useful/necessary especially when considering scenarios involving multiple inputs.

\subsection{Separate Laguerre time scale for each Laguerre series}
Earlier in this document the Laguerre time scale was discussed as an important parameter in the Laguerre polynomial. 

The Laguerre polynomial acts as a filter with time constant $1/a$. The time scale should be chosen to be the time constant of the response that is being modeled. Since a separate time scale can be chosen for each Volterra term this also allows adaptations to differences in responses within the same input. Again similar to the example above consider an input with a low frequency first order relationship to the output and a high frequency second order relationship to the output. A situation that would probably occur more frequently would be two inputs with significantly different dynamics.

\subsection{A Generalized Algorithm}
With these three modifications the equations given in the previous section need to be modified. Assume that there are $D$ I/O points being considered.

\begin{align}
\tilde{y}(k) &= [\mathbf{U}_{k}^{[1]},\mathbf{U}_{k}^{[2]},\ldots,\mathbf{U}_k^{[N]}]\mathbf{\Theta} \label{eq:MISOMatrixAPCONLI}
\end{align}
Where: 
\begin{align}
\mathbf{U}^{[n]} &= [\mathbf{U^{n}B^{n}}]^{[n]},  n=1,\ldots,N \\
\mathbf{U}^{n} &=  [\mathbf{u^1},\mathbf{u^2},\ldots,\mathbf{u^I}]
\end{align}

\begin{align}
\mathbf{u^{i}}&=
\begin{bmatrix}
u_i(k) &u_i(k-1) &u_i(k-2) &\cdots &u_i(k-M)   \\
u_i(k+1) &u_i(k) &u_i(k-1) &\ddots &\ddots \\
u_i(k+2) &u_i(k+1) &u_i(k) &\ddots &\ddots \\
\vdots &\ddots &\ddots &\ddots &\ddots \\
u_i(k+D) &u_i(k+D-1) &\cdots &\ddots &u_i(k+D-M)
\end{bmatrix}
, i=1,\ldots,I
\end{align}

\begin{align}
\mathbf{B^{n}} &= 
\begin{bmatrix}
\mathbf{B_{1}^{n}} &\mathbf{0} &\cdots &\mathbf{0}  \\
\mathbf{0} &\mathbf{B_{2}^{n}} &\ddots &\vdots   \\
\vdots &\ddots &\ddots &\mathbf{0}  \\
\mathbf{0} &\cdots &\mathbf{0} &\mathbf{B_{I}^{n}} 
\end{bmatrix}
\end{align}

\begin{align}
\mathbf{B_{i}^{n}} &= 
\begin{bmatrix}
l_1^{a_{n,i}}(0) & l_2^{a_{n,i}}(0) &\cdots &l_{R_{n,i}}^{a_{n,i}}(0) \\
l_1^{a_{n,i}}(1) & l_2^{a_{n,i}}(1) &\cdots &l_{R_{n,i}}^{a_{n,i}}(1) \\
\vdots &\vdots &\ddots &\vdots \\
l_1^{a_{n,i}}(M) & l_2^{a_{n,i}}(M) &\cdots &l_{R_{n,i}}^{a_{n,i}}(M) \\
\end{bmatrix}
\end{align}
Where:	$i=1,\ldots,I$ and $n =1,\ldots,N$. Finally:
\begin{align}
\mathbf{\Theta} = [\theta_{1,1,1},&\theta_{1,1,2},\cdots,\theta_{1,1,R_{n,i}}, \nonumber\\
&\theta_{1,2,1},\cdots,\theta_{1,I,R_{n,i}},\cdots,\theta_{N,I,R_{N,i}}]^T
\end{align}

The explanations and definitions of the following symbols should be noted. Recall that $a^{[n]}$ is the $n^{th}$ reduced Kronecker product. $N$ is the maximum Volterra order of the inputs $N=max(N_i)$. $I$ is the number of inputs. $k$ denotes the time step at which identification will begin on the data set, while this can be anywhere in the data set such that $k-M > 1$ usually $k=M+1$. $\mathbf{B^n}$ is a block diagonal matrix, the boldface zeros represent zero matrices with appropriate dimensions. $a_{n,i}$ is the Laguerre time scale that pertains to Volterra term $n$ and input $i$. These values can be specified or can be calculated by optimization of an initial guess, recall that this is a global optimization and that local minima may be a problem. $R_{n,i}$ is the order of the Laguerre polynomial that will be used to fit each polynomial, these values are specified by the user. Finally $\theta_{n,i,r}$ is the $\theta$ (or Laguerre coefficient) corresponding to Volterra term $n$, input $i$, and Laguerre polynomial $r$.

\subsection{Effect of Laguerre Reduction}\label{sec:VoltLagParameterization}
The parameterization of the Volterra series was discussed earlier in this document. It was identified to be a significant impediment to the practical application of Volterra model identification because of the number of I/O that are required to confidently identify such a large number of model parameters.
The replacement of the Volterra kernel with the Laguerre polynomials allows a reduction of the required parameters. This occurs because the Laguerre functions approximate each Volterra kernel with $R$ parameters instead of $M$ parameters (assuming R is the same for each term). The total number of coefficients in the reduced Volterra model $V_{(N,R)}$ can be derived similarly to \cref{eq:VoltParam} and is shown below in \cref{eq:VoltLagParam}.

\begin{align}
C_{(N,R)}=R^N \label{eq:VoltLagParam}
\end{align}

\autoref{tab:VoltLagParameterization} below shows the amount of parameters for a Volterra-Laguerre model with $N$ Volterra terms and an $R^{th}$ order Laguerre polynomial.

\begin{table}[h]
\centering
\begin{tabular}{|c|c|c|c|c|}
\hline
   & N=1 & N=2 & N=3 & N=4 \\ \hline
R=1 & 1   &  1  &  1  &  1  \\ \hline
R=2 & 2   &  4  &  8  &  16 \\ \hline
R=3 & 3   &  9	&  27 &  81 \\ \hline
R=4 & 4   &  16 &  64 &  256\\ \hline
\end{tabular}
\caption{Number of Volterra-Laguerre parameters based on N and R}
\label{tab:VoltLagParameterization}
\end{table}

The number of required parameters for a Volterra-Laguerre model is much less than that of a Volterra model. For example, consider a Volterra system with $M=50$ and $N=3$. \autoref{tab:VolterraParameterization} shows that this would require approximately $8000$ parameters to describe. Fitting the same Volterra model with $N=3$ and Laguerre polynomials of order $R=3$. The number of parameters is reduced to $27$ which is less than $0.5$ percent of the number required for the Volterra series.

It is important to remember that reduction of the number of required model parameters is desirable because of the amount of required I/O data to identify them. For a fixed amount of I/O data the identification of the parameters will have some statistical confidence inversely proportional to the number of parameters. If the number of required parameters are decreased then the statistical confidence will go up. For industrial applications this means that less time can be spent collecting data to achieve the same (or better) statistical confidence in the model parameters. The other way of looking at it is that a better model can be made with the same amount of I/O data. Either way use of the Volterra-Laguerre series can be extremely beneficial.

\subsubsection{Example - Variable Parameters}
A short demonstration highlights the value of having an algorithm that allows variation to $N$, $l_{r}$, and $a_{n,i}$. Data for this example was borrowed from \citet{bachlin2010wearable}. The dataset is a multivariate time-series for freezing of gait in patients with Parkinson's disease. 'Trunk acceleration - horizontal forward acceleration [mg]' and 'Upper leg (thigh) acceleration - horizontal forward acceleration [mg]' were used $u^{1}(t)$ and $u^{2}(t)$ respectively. And, 'Ankle (shank) acceleration - horizontal forward acceleration [mg]' was $y(t)$ . Hundreds of simulations were run using samples drawn uniformly from the following domains: $N$, the nonlinear degree or Volterra order was taken from integers $[1,5]$ for each input, $l_{r}$ was drawn from integers $[2,4]$ for each volterra kernel, and $a_{n,i}$ was taken from the set of real numbers on $[0.005, 100]$ for each laguerre series. Hundreds more simulations were run using equal $N$, $l_{r}$, and $a_{n,i}$ as well.

Compiling the results from these simulations and plotting the corresponding distributions of the sum of the squared error (SSE) indicates the difference in the expected error from a scenario with $N$, $l_{r}$, and $a_{n,i}$ fixed (according to practice in current literature) as opposed to the case where they can be different.

\begin{figure}
	\centering
		\includegraphics[width=5in]{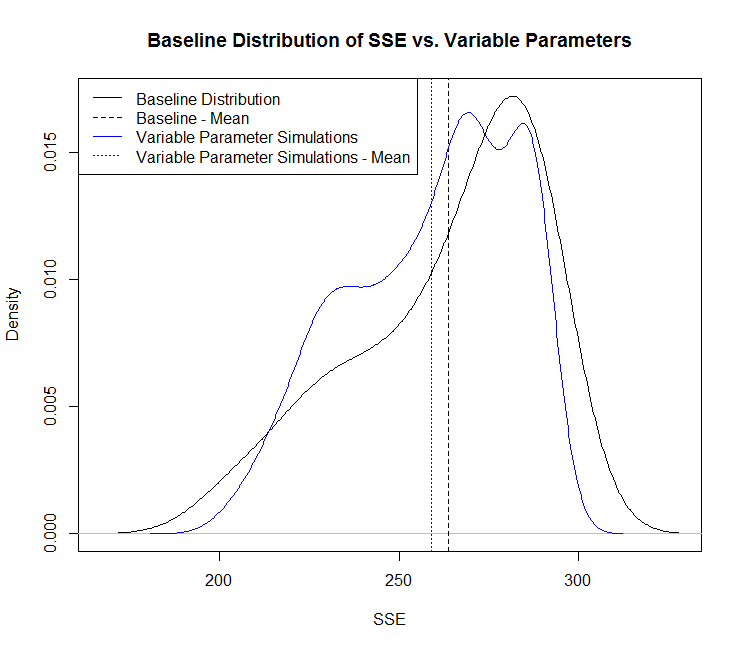}
		\caption{Distribution density of baseline simulations vs. variable parameter distributions.}
	\label{fig:SimulationDistribution}
\end{figure}

Normalizing the means to the lowest value of SSE the difference between the baseline simulations and the variable parameter simulations is approximately $5.5$ percent. It should be noted that the simulations with variable parameters have a tighter distribution. It should also be noted that lower SSE is not necessarily desirable because of over-fitting. This example illustrates that, in general a VL model fit with non-equal parameters will have less error than a model with equal parameters.

\section{Conclusion}
This paper presented the general Volterra series and discussed some typical simplifying assumptions. Limitations of the finite $V_{(N,M)}$ class Volterra Series were discussed with respect to physical systems. These limitations can often be beneficial as they represent behaviors not commonly seen in physical systems. Perhaps the key limitation to the Volterra series is the number of parameters required to describe a model, Laguerre Polynomials were introduced as a tool, by which to estimate the Volterra kernels and reduce the total number of model parameters required for a given system. A novel algorithm is presented to generalize identification of discrete VL systems. The proposed algorithm allows flexibility of parameterization to fit any class of system that can be described by a $V_{(N,M)}, l_{r}^{a_{n,i}}$ Volterra-Laguerre model. An example using the proposed algorithm on experimental data showed that using variable parameters has a higher probability of fitting the data set with less error. 


\newpage

\appendix
\section*{Appendix A. Memory}\label{sec:Memory}
\emph{Fading Memory} systems are those where the output exhibits a finite, steady state, response to a step input. This could be the velocity of a car due to an increase of fuel being fed to the engine. Or the increase in flow due to a change in valve position. 

Fading Memory and Finite Memory and synonymous. Any member of the class of \emph{fading memory} systems can be approximated to arbitrary accuracy by a finite Volterra model \citep{boyd1985fading} . A better discussion of \emph{fading memory} systems and the work done on them can be found in \citet[p. 41]{DoyleOgunnaikePearson2001},and \citet[Sec. III]{boyd1985fading}. It turns out that the concept of \emph{fading memory} has been around at least as long as the Volterra series itself. Regularly the Horizon should be equal to the number of steps in the time to steady state. In EHPC\index{EHPC} there has been work suggesting criteria for choosing the memory length or ``Horizon" of finite memory systems, see \citet{kong1994criteria} for more details and references.

An example of infinite memory is an integrating process such as the level of tank with respect to influent flow. If the influent flow rate of a tank steps from $0$ to a positive value of $x$ the tank will begin to fill. If the model ever ``forgets" that the flow rate was changed to $x$ then it would predict that the tank level should stop changing. Thus this model requires infinite memory to correctly predict the level of the tank.

Some more discussion regarding systems with infinite memory can be found in \citet[p.334]{Schetzen1980}. Systems with infinite memory can still be handled but require some special treatment. In the case of the above example the level signal could be differentiated giving a constant rate of increase of the level.

\section*{Appendix B. Orthogonal and Orthonormal Functions}\label{sec:Orthogonality}
Two vectors are orthogonal if they are perpendicular. In order to test if two vectors are perpendicular one can take the \emph{inner product} of the vectors, if they are perpendicular the inner product will be zero. In Euclidean space (i.e. x,y,z) the inner product is the \emph{dot product}. If the vectors are perpendicular their inner product will be zero. Non-zero orthogonal vectors are always \emph{linearly independent} which means that one of them can't be written as a combination of any finite combination of the others.

This idea of orthogonality can be extended to functions. In other words two functions are orthogonal if their inner product is zero. An \emph{orthogonal set} is a group of vectors that are orthogonal to each other. For orthogonal functions the orthogonality condition can be expressed as \cref{eq:OrthogCond} below.

\begin{align}
\int_{a}^{b} w_m(x)w_n(x)dx= \label{eq:OrthogCond}
\begin{cases}
\lambda_n & \text{for } m=n \\
0 & \text{for } m\neq n
\end{cases}
\end{align}

Here,$w_n(x)$ is an orthogonal set of functions over the interval $[a,b]$. $\lambda_n$ is the product of $w_n$ with itself and is therefore always positive.

A function $f(x)$ can be approximated in the interval $[a,b]$ by $N$ members of the orthogonal set, yielding \cref{eq:OrthogApprox} below. Here $c_n$ are coefficients chosen to minimize the error between the left hand side of \cref{eq:OrthogApprox} and the right hand side.

\begin{align}
f(x)\approx \sum_{n=1}^{N} c_nw_n(x) \label{eq:OrthogApprox}
\end{align}

The equation for error can be represented as \cref{eq:err} below. This is then squared to give \cref{eq:ISE} . Substituting \eqref{eq:err} into \eqref{eq:ISE}  yields \cref{eq:ISEsimp} . Which will only be finite if \eqref{eq:ISErestrict} is true.

\begin{align}
e_N(x)&=f(x)-\sum_{n=1}^{N} c_nw_n(x) \label{eq:err} \\
I_N&=\int_{a}^{b}e_n^2(x)dx \label{eq:ISE} \\
I_N&=\int_{a}^{b} \left[f(x)-\sum_{n=1}^{N} c_nw_n(x)\right]^2 dx \label{eq:ISEsimp} 
\end{align}
\begin{equation}
\int_{a}^{b} f^2(x) dx < \infty \label{eq:ISErestrict}
\end{equation}

It can be shown that for all $N$ \cref{eq:I_Nmin} holds. If the orthogonal set is complete (i.e. $N=\infty$) \cref{eq:I_Nmininfty} holds (See \citet[Sec. 9.2]{Schetzen1980}).

\begin{align}
\sum_{n=1}^{N} c_n^2\lambda_n &\leq \int_{a}^{b}f^2(x) dx \label{eq:I_Nmin} \\
\sum_{n=1}^{\infty} c_n^2\lambda_n &= \int_{a}^{b}f^2(x) dx \label{eq:I_Nmininfty}
\end{align}

An orthogonal set is \emph{orthonormal} if the magnitude of $\lambda_n$ in \cref{eq:OrthogCond} is equal to $1$ for all values of $n$. Or \cref{eq:OrthogCond} can be rewritten as \cref{eq:OrthonormCond} below.

\begin{align}
\int_{a}^{b} w_m(x)w_n(x)dx= \delta_{mn} \label{eq:OrthonormCond}
\begin{cases}
1 & \text{for } m=n \\
0 & \text{for} m\neq n
\end{cases}
\end{align}

To satisfy \cref{eq:OrthonormCond} it is sufficient to meet the requirements of \cref{eq:OrthonormCond_a,,eq:OrthonormCond_b}. This derivation can be found in more detail in \citet[Sec. 9.2 and 16.1]{Schetzen1980}.

\begin{equation}
\int_{a}^{b} w_m(x)w_n(x)dx= 0 \text{, for } m<n \label{eq:OrthonormCond_a}
\end{equation}
\begin{equation}
\int_{a}^{b} w_m^2(x)dx=1 \label{eq:OrthonormCond_b}
\end{equation}

\section*{Appendix C. Laguerre Polynomials - Properties and Useful Forms}\label{sec:LaguerreProps}
\Cref{eq:GeneralLaguerre} can also be represented in the following form:

\begin{align}
l_{n}(t) &= \mathpzc{L}_{n}(t)e^{-at}\label{eq:GenLaguerrePoly}\\
\mathpzc{L}_{n}(t)&=
 \sqrt{2a}\sum_{k=0}^{n}\frac{(-1)^{k}n!2^{n-k}}{k![(n-k)!]^2}(2at)^{n-k}
\end{align}

The $n^{th}$ degree polynomial $\mathpzc{L}_{n}(t)$ is called the $n^{th}$
Laguerre polynomial. It is also interesting to note that the Laguerre function
$l_{n}(t)$ has $n$ zero crossings defined by the zeros of $\mathpzc{L}_{n}(t)$.
 More detail concerning the derivation and properties of the Laguerre polynomials and functions can be found in \citet[Ch.~16]{Schetzen1980} and \citet[Sec.~18.5]{CommunicationLee1960}.

\vskip 0.2in
\bibliography{JMLR_VolterraLaguerre}

\end{document}